\documentclass[letterpaper, 10 pt, conference]{./ieeeconf}

\IEEEoverridecommandlockouts    
\usepackage[utf8]{inputenc}
\usepackage[T1]{fontenc}

\usepackage{graphics} 
\usepackage{epsfig} 
\usepackage{amsmath} 
\DeclareMathOperator*{\argmax}{arg\,max}

\usepackage{amssymb}  
\usepackage{caption}
\usepackage{subcaption}
\usepackage{xcolor}
\usepackage{hyperref}
\usepackage{algorithm2e}
\usepackage{multirow}
\usepackage{float}
\RestyleAlgo{ruled}

\title{\LARGE \bf Adaptive Risk-Tendency: Nano Drone Navigation in Cluttered Environments with Distributional Reinforcement Learning}

\author{Cheng Liu$^1$, Erik-Jan van Kampen$^1$, Guido C.H.E. de Croon$^1$
\thanks{$^1$ Delft University of Technology, Email: c.liu-10@tudelft.nl, e.vankampen@tudelft.nl, g.c.h.e.decroon@tudelft.nl.}}

\begin{document}

\maketitle
\thispagestyle{empty}
\pagestyle{empty}

\begin{abstract}
Enabling the capability of assessing risk and making risk-aware decisions is essential to applying reinforcement learning to safety-critical robots like drones. In this paper, we investigate a specific case where a nano quadcopter robot learns to navigate an apriori-unknown cluttered environment under partial observability. We present a distributional reinforcement learning framework to generate adaptive risk-tendency policies. Specifically, we propose to use lower tail conditional variance of the learnt return distribution as intrinsic uncertainty estimation, and use exponentially weighted average forecasting (EWAF) to adapt the risk-tendency in accordance with the estimated uncertainty. In simulation and real-world empirical results, we show that (1) the most effective risk-tendency vary across states, (2) the agent with adaptive risk-tendency achieves superior performance compared to risk-neutral policy or risk-averse policy baselines. Code and video and can be found in this repository: \url{https://github.com/tudelft/risk-sensitive-rl.git}
\end{abstract}

\section{Introduction}

Reinforcement learning (RL) is promising in solving sequential decision making problems such as robotic navigation with obstacle avoidance as it seeks long-term optimal policies \cite{RL1998Sutton, Q-learningML1992}. Recent advances in deep reinforcement learning (DRL), which combines deep neutral networks with RL, have shown the capability of achieving super human performance in diverse complex environments \cite{natureDQN2015Mnih, DDPG2016ICLR, EndToEndVisuomotor2015Levine}. The majority of current DRL methods are designed for maximizing the expectation of accumulated future returns, omitting to consider the risk of rare catastrophic events. However, when it comes to applying RL to safety-critical robots like drones, instead of aiming at achieving a high expected return, dealing with risks and making decisions under uncertainty is crucial and remains a challenge.

A natural way to risk-sensitive RL is considering the worst-case of the stochastic return rather than its expectation, but this may lead to over-conservative policies \cite{HowRoRisk}. Recent works proposed to model the distribution of the future return and to generate multiple policies with different risk-sensitivities by changing levels of a risk metric \cite{TangZS19}. While \cite{TangZS19} captures the stochasticity in accumulated returns by approximating the mean and variance of a gaussian distribution, \textit{distributional RL} reconstructs the true intrinsic distribution of future returns \cite{C51,QR,IQN2018}. A major merit of distributional RL is that it can generate multiple policies with different levels of risk-tendencies \cite{RAAC, IQN2018, DSAC}.

\begin{figure}
    \centerline{\includegraphics[width=0.5\textwidth]{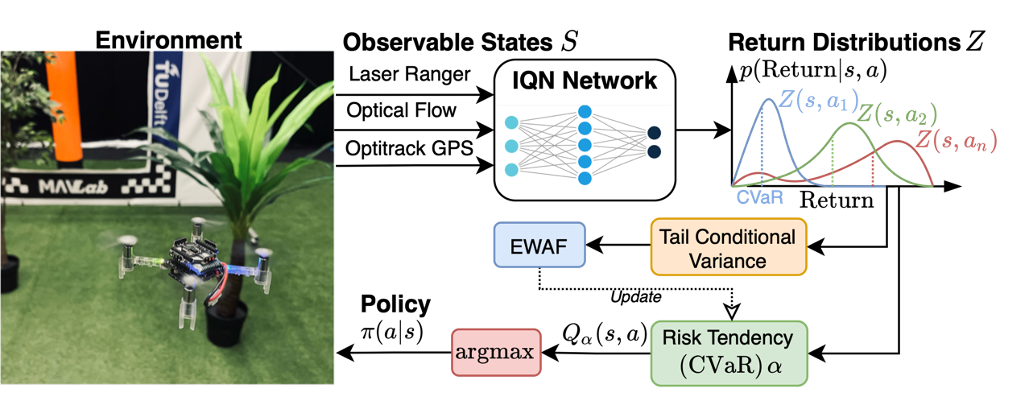}}
    \caption{ART-IQN framework that enables a Crazyflie \cite{Crazyflie} nano drone navigating through a cluttered environment under partial observability with adaptive risk-tendency.}
    \label{fig:drone_demo}
\end{figure}

Distributional RL has been applied to safety-critical applications such as autonomous driving at occluded intersections \cite{MinimizeCar} and mobile-robot indoor navigation \cite{DSAC-condition}. These methods learn a policy that can vary its risk-tendency during training, but they still rely on a fixed risk-tendency for each deployment task. However, a proficient pilot would not be as cautious while cruising in fair weather as when landing in stormy weather. In other words, the ideal degree of risk-tendency varies as a function of not only the task but also the real-time feedback from the environment. A step towards building intelligent robots is \textit{adapting risk-tendency} on the fly automatically.

To achieve this goal, we propose Adaptive Risk-Tendency Implicit Quantile Network (ART-IQN) that can adapt risk-tendency by reacting to the context. We propose to let intrinsic uncertainty \cite{uncertainty} (estimated by lower tail conditional variance) set the way in which the agent acts - adapting risk-tendency by forecasting the intrinsic uncertainty.
The effectiveness of ART-IQN is validated on safety-critical tasks - autonomous drone navigation in cluttered environments with constrained sensors (shown in Fig. \ref{fig:drone_demo}). Both in simulation and real-world experiments, our method shows superior performance in the trade-off between navigation efficiency and safety in comparison with risk-neutral and risk-averse baselines. Our main contributions are:

\begin{itemize}
    \item an automatic adaption in risk-tendency on the fly in accordance with intrinsic uncertainty estimation;
    \item a drone navigation algorithm based on distributional RL, that can learn a variety of risk-sensitive policies;
    \item a sim-to-real RL framework and a light-weight simulation environment, enabling seamless transfer from simulation to reality.
\end{itemize}

\section{Related Work}

\subsection{Risk and Uncertainty in RL-based Navigation}
RL-based robot navigation methods surged recently due to the capability of generalization and robustness\cite{DRL2018introduction, mobileRobotNavMapless2017iros}. Several navigation and obstacle avoidance algorithms based on RL have also emerged to address risks and uncertainties in the environment. For instance, \cite{uncertaintyAware-MPC-RL-2017} proposed to use a neural network to predict collision probability at future steps for obstacle avoidance tasks, utilizing MC-dropout \cite{MC-dropout2016Y.Gal} and bootstrapping \cite{BootstrappedDQN} to estimate the uncertainty of the model prediction. Additionally, \cite{SafeRLWithModelUncertaintyEstimation} enabled estimation of the regional increase of uncertainty in novel dynamic scenarios by introducing LSTM \cite{LSTM} to add memory of historical motion of the robot. \cite{ResillientBehaviorForNavigation} resorts to a model-free policy network as action selector, a GRU \cite{GRU} to predict uncertainty in local observation and uses the prediction variance to adjust the variance of the stochastic policy.

However, these methods either use MPC \cite{MPC} as action selector, which consumes a lot of computational resources, or they require an additional predictor model to estimate the uncertainty. In our method, the risk measure and uncertainty estimation are easily and efficiently implemented based on distributional RL (more details in Section \ref{sec:method}), which requires minimal additional computational resources.

\subsection{Distributional Reinforcement Learning}
\label{sec:drl}
Distributional RL has gained momentum recently, which takes into account the whole distribution of value functions rather than the expectation \cite{TangZS19, C51, QR, IQN2018}. Since the whole distribution contains more information beyond the first moment, one can utilize it to make more informed decisions that lead to higher rewards. Recent literature shows that similar mechanisms also exist in human brains \cite{naturedistri}. 

A forerunner of distributional RL is categorical DQN \cite{C51}, which uses categorical distribution with fixed supports to approximate the probability density function (PDF) of the return. A more flexible way to approximate the distribution is quantile regression \cite{quantileRegressionModels1998}. For instance, the quantile regression DQN (QR-DQN) algorithm \cite{QR} learns the distribution by approximating the quantile function (QF) with fixed quantile values. The implicit quantile network (IQN) algorithm \cite{IQN2018} further improved the flexibility and approximation accuracy compared to QR-DQN by learning quantile values from quantile fractions sampled from a uniform distribution $\mathcal{U}[0, 1]$. This is achieved with a deep neutral network representing the QF by mapping quantile fractions to quantile values under Wasserstein distance, a loss metric which indicates the minimal cost for transporting mass to make two distributions identical \cite{IQN2018}.

There are also applications of distributional RL to safety-critical environments in the literature. \cite{MinimizeCar} incorporates IQN to solve an autonomous driving task at intersections by combing risk-averse IQN with safety guarantees. Based on \cite{DSAC}, \cite{DSAC-condition} proposes a method enabling a mobile robot navigating office scenarios with multiple risk-sensitivities.

Even though risk-tendency can be altered without retraining a policy, those methods require a fixed risk-tendency for each deployment task.
Our algorithm is able to adjust its risk-tendency by reacting to dynamic uncertainty levels rather than following a fixed manually set risk-tendency.

\section{Methodology}
\label{sec:method}

\subsection{Problem Statement}
\label{sec:method-a}
We formulate the drone navigation task as Partially Observable Markov Decision Process (POMDP) \cite{spaan2012partially}.
\subsubsection{POMDP Setup}
The POMDP can be defined as a tuple $(\mathcal{S}, \mathcal{A}, \mathcal{O}, \mathcal{P}, R, \gamma)$, where $\mathcal{S}$, $\mathcal{A}$ and $\mathcal{O}$ represent the state, action and observation spaces. The drone interacts with the environment in discrete timesteps. At each timestep $t$, it receives the observation $o_t \in \mathcal{O}$ from the environment and performs an action $a_t \in \mathcal{A}$ based on its policy function $\pi_t(a_t | o_t)$, which causes a transition of the state from $s_t$ to $s_{t+1} \sim \mathcal{P}(\cdot | s_t, a_t)$, generating a reward $r_t = R(s_t, a_t)$ and a new observation $o_{t+1} \sim \mathcal{O}(\cdot|s_{t+1}, a_t)$. Following policy $\pi$, the discounted sum of future rewards is denoted by the random variable $Z^{\pi}(s_t, a_t) = \sum_{k=0}^{\infty}\gamma^{k}R(s_{t+k}, a_{t+k})$ with $\gamma \in (0, 1)$ as the discount factor. Standard RL aims at maximizing the expectation of $Z^{\pi}$, which is known as the action-value function $Q^\pi(s_t,a_t) = \mathbb{E}[Z^\pi(s_t, a_t)]$.

\begin{figure}[!h]
    \centering
    \includegraphics[width=0.43\textwidth]{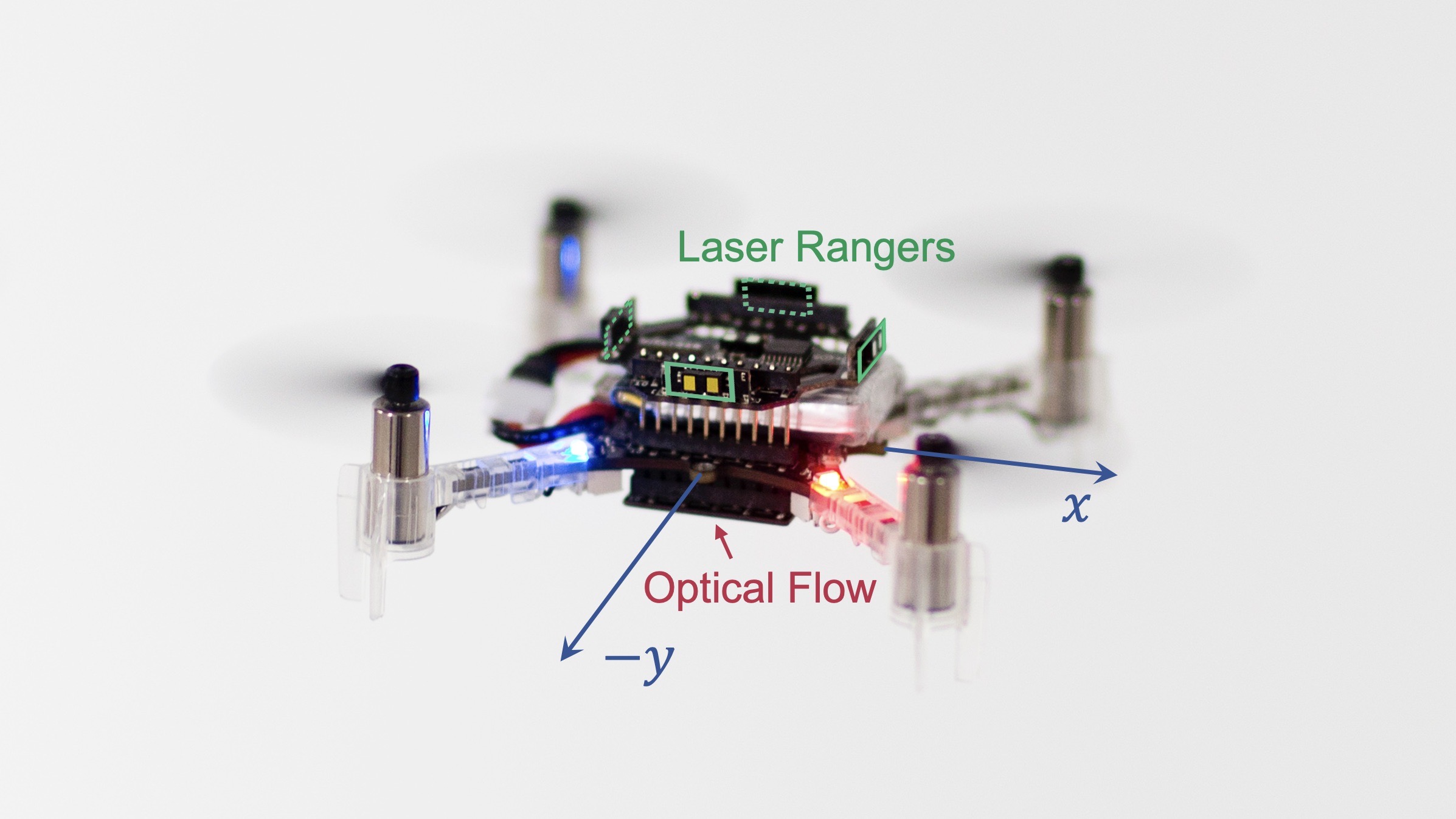}
    \caption{A Crazyflie with 4 lasers to detect obstacles. [Picture by Guus Schoonewille, reprinted TU Delft]}
    \label{fig:crazyflie}
\end{figure}

\subsubsection{States and Observations}
\label{sec:space}

We use Crazyflie nano quadrotor as our experiment platform. As shown in Fig. \ref{fig:crazyflie}, the Crazyflie is equipped with four lasers in the drone's positive and negative $x$ and $y$ axis to detect obstacles. It also has an optical flow camera to estimate velocity for low-level flight control. Given a navigation task, $\mathcal{S}$ contains information about the drone itself, the goal and obstacles. The state can be parameterized as $s_t = \langle\mathbf{p}, d_g, \mathbf{d}_{o}\rangle$, where $\mathbf{p}$ is drone's global position, $d_g = ||\mathbf{p} - \mathbf{p}_g||_2$ is the distance from the drone to the goal, and $\mathbf{d}_{o}$ is a vector consisting of distances from the drone to surrounding obstacles.

Due to the constrained sensors onboard, $\mathcal{S}$ is not fully observable to the drone. Instead, the drone receives a partial observation which is formulated as a tuple $o_t = \langle\mathbf{p}, d_g, \mathbf{d}_{l}\rangle$ where $\mathbf{d}_{l}$ denotes laser reflections. The laser detects obstacles at a maximum range of $4$ meters. $\mathbf{p}$ and $d_g$ are given by a global motion capturing system in real-world experiments.

\subsubsection{Action Space}
To incorporate our algorithm, $\mathcal{A}$ consists of discretized velocities with multiple magnitudes and directions.
The velocity magnitudes that can be chosen are $m$ discretized values exponentially spaced in $(0, v_{m}]$, in which $v_{m}$ is the maximum velocity. Since only obstacles that intersect with the laser beams will be detected, there are 4 possible moving directions evenly spaced between $[0, 2\pi)$.

\subsubsection{Reward Function}
The reward function is manually designed to award the drone for reaching the goal as fast as possible, while penalizing for collisions or getting close to obstacles:

\begin{equation}
    R(s_t, a_t) = \left\{
    \begin{array}{lll}
    50                          &  d_g < d_f &\\
    5(d_o-d_s)            &  r_d < d_o < d_{s} & \\
    -25                         &  d_o < r_d & \\
    -0.1 &\text{otherwise,} \\
    \end{array}\right.
\end{equation}
where $d_f$ is the goal-reaching threshold, ${d}_{o}$ is the distance from drone to the closest obstacle, $d_s$ is the safety margin, $r_d$ represents the radius of drone.

\subsection{Adaptive Risk-Tendency Implicit Quantile Network}

To adjust the risk-tendency on the fly dynamically, we propose the Adaptive Risk Tendency Implicit Quantile Network (ART-IQN) algorithm. We introduce the key components of ART-IQN as shown in Fig. \ref{fig:drone_demo}, which are (1) the risk-sensitive IQN, (2) the intrinsic uncertainty estimation and (3) the EWAF uncertainty forecasting.

\subsubsection{Implicit Quantile Network}
\label{sec:iqn}
In distributional RL, the distributional Bellman equation \cite{C51} can be defined as

\begin{equation}
    Z^\pi(s, a) \stackrel{D}{=}R(s, a) + \gamma Z^\pi(s', a'),
    \label{eq:bellman}
\end{equation}
where $\stackrel{D}{=}$ denotes equality in distribution, state $s'$ and action $a'$ at next timestep are distributed according to $s'\sim \mathcal{P}(s,a), a' \sim \pi(\cdot|s')$.

We represent $Z^\pi(s, a)$ implicitly by its quantile function as in IQN \cite{IQN2018}. Concretely, the quantile function is approximated by a neural network with learnable parameters $\theta$. We express such implicit quantile function as $Z_{\theta}^\pi(s,a;\tau)$, where $\tau \in [0, 1]$ is the quantile level. To optimize $\theta$, quantile regression \cite{quantileRegressionModels1998} is used with quantile Huber-loss as a surrogate of the Wasserstein distance \cite{IQN2018}.

A neural network with parameters $\theta'$ is used as the target distribution approximator and the temporal difference (TD) at sample $(s,a,r,s')$ is computed as

\begin{equation}
    \delta_{\tau, \tau'} = r + \gamma Z_{\theta'}^\pi(s',a';\tau') - Z_{\theta}^\pi(s,a;\tau),
\end{equation}
for $\tau$, $\tau'$ independently sampled from the uniform distribution, i.e. $\tau, \tau' \sim \mathcal{U}[0,1]$.

The $\tau$-quantile Huber-loss is defined as
\begin{equation}
\centering
    \begin{aligned}
    & \rho_\kappa(\delta; \tau) = |\tau-\mathbb{I}\{\delta < 0\}|\frac{\mathcal{L}_\kappa(\delta)}{\kappa}, \text{with} &\\
    & \mathcal{L}_{\kappa}(\delta) = \left\{
    \begin{array}{ll}
    \frac{1}{2}\delta^2  & \text{if} |\delta|\leq \kappa\\
    \kappa(|\delta| - \frac{1}{2}\kappa)  & \text{otherwise,} \\
    \end{array}\right. &
    \end{aligned}
\end{equation}
where $\mathbb{I}$ is an indicator operator. The threshold $\kappa$ provides smooth gradient-clipping. We approximate the quantile loss by sampling $N$ independent quantiles $\tau$ and $N'$ independent targets $\tau'$. The loss function to update $\theta$ is

\begin{equation}
\label{eq:loss}
    \mathcal{L}(\theta) = \frac{1}{N\cdot N'}\sum\limits_{i=1}^{N}\sum\limits_{j=1}^{N'}\rho_{\kappa}(\delta_{{\tau_i}, {\tau'_j}};\tau_i).
\end{equation}

By backpropagating $\mathcal{L}(\theta)$ with respect to $\theta$, the Wasserstein distance is minimized between the current return distribution $Z^\pi(s, a)$ and the target $R(s, a) + \gamma Z^\pi(s', a')$.

\subsubsection{Risk-sensitive Policy and Risk Metric}
\label{sec:cvar}
Distributional RL is inherently risk-sensitive by combining \textit{risk metrics} \cite{HowRoRisk} to create a \textit{distorted expectation} \cite{DSAC} on the return distribution.
A distorted expectation is a risk weighted expectation of the distribution under a specific distortion function - which indicates a non-decreasing function $\beta:[0,1]\rightarrow [0, 1]$ satisfying $\beta(0) = 0$ and $\beta(1) = 1$. The distorted expectation of $Z$ under $\beta$ is defined as $Q_\beta = \int_0^1F_Z^{-1}(\tau)d\beta(\tau)$, where $F_Z^{-1}(\tau)$ is the quantile function or cumulative density function. According to \cite{IQN2018}, any distorted expectation can be represented as a weighted sum over the quantiles. A corresponding sample-based risk-sensitive policy is obtained by approximating $Q_\beta$ by $K$ samples of $\tilde{\tau} \sim \mathcal{U}[0, 1]$:




\begin{equation}
\label{eq:policy}
    \pi_{\beta}(s) = \argmax_{a \in \mathcal{A}}\frac{1}{K}\sum\limits_{k=1}^{K}Z_{\beta(\tilde{\tau}_k)}(s,a).
\end{equation}

Altering the sampling principle for $\tau$ creates various risk-sensitive policies. Specifically, we consider Conditional Value-at-Risk (CVaR) \cite{CVarFinance}, a \textit{coherent risk metric} \cite{HowRoRisk} as our distortion function.
CVaR is applied to IQN by modifying $\tilde{\tau} \sim \mathcal{U}[0, 1]$ to $\tilde{\tau} \sim \mathcal{U}[0, \alpha]$, where $\alpha$ is the CVaR value. We get risk-averse policies as $\alpha$ decreases to near zero and reduce back to risk-neutral when $\alpha=1$.

\subsubsection{Lower Tail Conditional Variance for Intrinsic Uncertainty Estimation}
\label{sec:rtv}
One major source of risk comes from intrinsic uncertainty, which is due to stochasticity of the environment or partial observability. Opposite to \textit{epistemic uncertainty} \cite{uncertainty}, intrinsic uncertainty is independent of the agent's knowledge about the task.

In distributional RL, a more or less spread out return distribution acts as a measure of intrinsic uncertainty \cite{Tactical}.
Inspired by \cite{explordistri}, where the decaying upper tail conditional variance of the return distribution is used for more efficient exploration, we use the lower tail conditional variance as the intrinsic uncertainty estimation for risk-tendency adaption. The lower half tail conditional variance is equivalent to the \textit{right truncated variance} (RTV):

\begin{equation}
\label{eq:rtv}
    \text{RTV} = \frac{2}{N}\sum\limits_{i=1}^{\frac{N}{2}}(F_Z^{-1}(\tau_i) - F_Z^{-1}(\tau_{\frac{N}{2}}))^2,
\end{equation}
in which $\tau_i$ are $\frac{i}{N}$-th quantile levels. Intuitively, RTV is biased towards negative returns.
We calculate RTV with respect to the median rather than the mean due to its statistical robustness \cite{explordistri, Huber.Wiley.ea1981Robuststatistics}. Note that $F_Z^{-1}(\tau)$ is implicitly approximated by $Z_{\theta}^\pi(;\tau)$ in our method.


\subsubsection{EWAF for Risk-Tendency Adaption}
\label{sec:ewaf}

In our framework, the risk-tendency can be cast in the choice of CVaR.
To formulate CVaR as a function of RTV, inspired by \cite{Tactical}, we propose to model CVaR by an exponentially weighted categorical distribution. Specifically, consider a categorical distribution $C$ with two logits: $C_i = \exp{(w_i)} / \sum_{i}{}\exp{(w_i)}$, in which $w_i \in \mathbb{R}, i = 1, 2$. By letting $\alpha = C_1$, the CVaR is restricted to a range of $(0, 1)$ and can be adjusted by altering the logit weights. Concretely, at each timestep $t$, the CVaR is adapted by updating $w_i$ with feedback $f$ by a step size $\eta$: $w_1 = w_1 - \eta f, w_2 = w_2 + \eta f$. We set $f = \text{RTV}_t - \text{RTV}_{t-1}$ as an indicator of intrinsic uncertainty feedback.
To avoid the CVaR approaching zero, an additional term: $\sum_{i}{}\exp(w_i) / b$ is added both to the denominator of $C_i$ for $i=1,2$ and to the numerator of $C_1$, which results in a CVaR range of $(\frac{1}{b+1}, 1)$. For example, $\alpha \in (0.1, 1)$ when $b=9$.

Up to now, we get ART-IQN that can adapt its risk-tendency by reacting to intrinsic uncertainty variations as explained in Algorithm \ref{alg:art-iqn}. In principle, when RTV is increasing and the current CVaR is relatively large, the agent will behave more risk-unwillingly by choosing smaller CVaR.

\SetKwComment{Comment}{\triangleright}{}
\SetAlgoNoLine%
    \begin{algorithm}[hbt!]
    \caption{ART-IQN for Drone Navigation}
    \label{alg:art-iqn}
    \SetKwInput{KwInput}{Input}
    \KwInput{Post-training $\texttt{IQN}_{\theta}$; $\mathcal{A}, K, N, w_1, w_2, b, \eta$}
    Initialize state $s$, CVaR $\alpha$\\
    \While{$d_g > d_f$ and $t \le H$}{
        Observe $o_t \leftarrow \langle\mathbf{p}, d_g, \mathbf{d}_{l}\rangle$  from $s_t$\\
        Get quantile function $Z_\theta(o_t, a; \tau) = \texttt{IQN}_{\theta}(o_t; \tau)$\\
        Distorted sampling $\tilde{\tau_k} \sim \mathcal{U}[0, \alpha], k = \{1, \ldots, K\}$\\
        Take action $a_t = \argmax\limits_{a \in \mathcal{A}}\frac{1}{K}\sum_{k=1}^{K}Z_\theta(o_t, a; \tilde{\tau_k})$\\
        Calculate right truncated variance: \\
        $\text{RTV}_t = \frac{2}{N}\sum_{i=1}^{\frac{N}{2}}(Z_\theta(o_t, a_t; \tau_i) - Z_\theta(o_t, a_t; \tau_{\frac{N}{2}}))^2$ \\
        Obtain feedback $f_t = \text{RTV}_t - \text{RTV}_{t-1}$\\
        Forecasting $w_{1} = w_1 - \eta f_t, w_{2} = w_2 + \eta f_t$\\
        Adapt CVaR $\alpha =  \frac{(b + 1)\exp{(w_1)} + \exp{(w_2)}}{(b+1)\sum_{i}{}\exp{(w_i) }}, i=1, 2$ \\
    }
    \end{algorithm}


\subsection{Training and Evaluation Pipelines}

\subsubsection{Environment}
\begin{figure}[!h]
    \centering
    \includegraphics[width=0.49\textwidth]{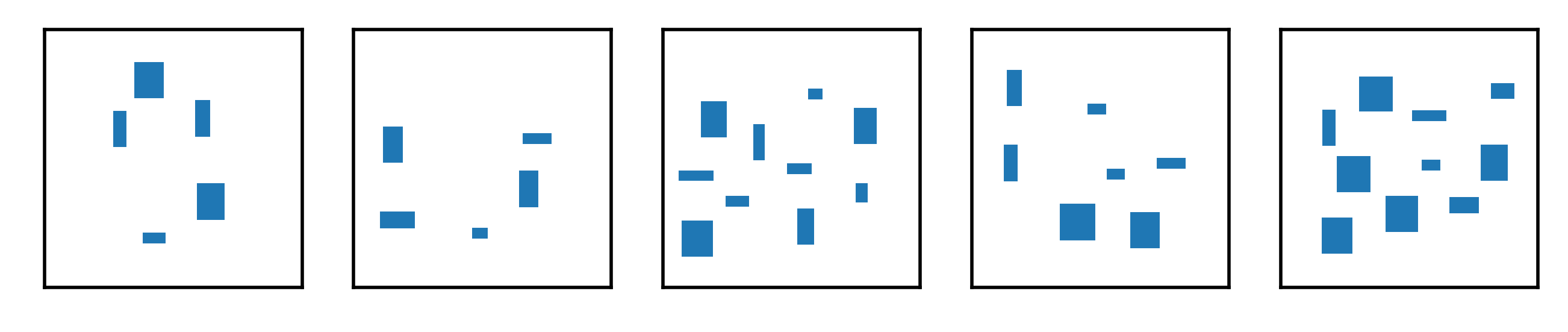}
    \caption{Randomly generated environments for training.}
    \label{fig:obstacle_demo}
\end{figure}

We design an OpenAI gym \cite{OpenAIgym} like $2$D environment for training with state, observation and action spaces defined in Section \ref{sec:space}. To achieve fast simulation, the drone is modelled as a point mass and the velocity command is assumed to be immediately executed. The state is updated every $T$ seconds in simulation time. We utilize \textit{domain randomization} \cite{domainRandom2017IROS} to train policies that can generalize to diverse scenarios. Specifically, the goal distance is uniformly sampled $d_g \sim \mathcal{U}[5, 7] (m)$ with drone initialized at a fixed start point for each training episode. The number of obstacles, the shape and position of each obstacle are randomly generated for each episode as demonstrated in Fig. \ref{fig:obstacle_demo}. Laser beams and obstacle outlines are modelled as line segments for simplicity. Gaussian noise $\mathcal{N}(\mu, \sigma)$ with mean $\mu=0.0$ and standard deviation $\sigma=0.01$ is added to the measurement of each laser to simulate a noisy sensor.

\subsubsection{Training Process}
We follow \textit{curriculum learning} to train the agent - the complexity of the environment increases as training process goes on. The first stage of training is implemented with relative small number of randomized obstacles, $n_{obs} \in [0, 5]$. After training several episodes (until a navigation success rate of $0.8$ is reached), the complexity of the environment is increased by adding more obstacles, $n_{obs} \in [6, 12]$. To make sure the agent accumulates a diverse range of experiences under a variety of risk-tendencies, the CVaR value is uniformly sampled $\alpha \sim \mathcal{U}(0, 1]$ at each episode. An episode is terminated after a collision or if the goal is not reached within $H$ timesteps. The whole training process is ended after the average return is empirically converged. It took $\approx 3.5$ hours on a $2.2$ GHz Intel i$7$ Core CPU, achieving a success rate of $0.88$.

The agent is modelled as a fully connected network that has $3$ hidden layers with $512$ units per layer. Each fully-connected layer is followed with a ReLU \cite{ReLU2018Agarap} activation function except the output layer. Adam \cite{Adam2015ICLR} is used as our optimizer with $lr$ as the learning rate. In each update step for every $D$ episodes, a batch size of $B$ samples are drawn from the experience replay buffer with size $E$. The hyper-parameters used are listed in Table \ref{tab:hyper}.


\begin{table}[!h]
\centering
    \caption{\uppercase{hyper-parameters}}
\begin{tabular}{cl|cl|cl|cl}
\hline
\multicolumn{8}{c}{Hyper-parameter symbols and values}  \\ 
\hline 
$lr$ & $2 \times 10^{-4}$   &  $v_{m}$ & 1 $[m/s]$   & $D$    & 5 &  $N, N'$  & 16  \\
$E$    & $5 \times 10^{4}$  & $r_d$    & 0.05 $[m]$  &  $K$   &   64 & $m$     & 3   \\ 
$\gamma$   & 0.99 & $d_f$    & 0.1 $[m]$   & $B$    & 32 & $b$   & 9 \\
$T$  & 0.1 [$s$]    & $d_s$    & 0.2 $[m]$   & $H$    & 200    &   $\eta$   & 0.5  \\ \hline

\end{tabular}
    \label{tab:hyper}
\end{table}

\begin{table*}[]
\centering
\caption{\uppercase{Quantitative Simulation Results}}
\label{tab:eval}
\begin{tabular}{c|c|ccc|ccc|ccc|ccc}
\hline
                                   & \multicolumn{1}{c|}{CVaR} & \multicolumn{3}{c|}{Average episodic return (mean $\pm$ std)}                                                                  & \multicolumn{3}{c|}{Success rate}                                                            & \multicolumn{3}{c|}{Collision rate}                                                          & \multicolumn{3}{|c}{Navigation time $[s]$}                                                     \\ \hline
\multicolumn{1}{c|}{$n_{obs}$} & \multicolumn{1}{c|}{-}    & \multicolumn{1}{c}{2}                         & \multicolumn{1}{c}{6}                          & 12                         & \multicolumn{1}{c}{2}             & \multicolumn{1}{c}{6}             & 12            & \multicolumn{1}{c}{2}             & \multicolumn{1}{c}{6}             & 12            & \multicolumn{1}{c}{2}             & \multicolumn{1}{c}{6}             & 12             \\ \hline 
DQN\cite{natureDQN2015Mnih}                               & \multicolumn{1}{c|}{-}    & \multicolumn{1}{c}{37.45 $\pm$ 9.01}          & \multicolumn{1}{c}{25.60 $\pm$ 13.91}          & 17.16 $\pm$ 15.85          & \multicolumn{1}{c}{0.85}          & \multicolumn{1}{c}{0.65}          & 0.56          & \multicolumn{1}{c}{0.13}          & \multicolumn{1}{c}{0.31}          & 0.39          & \multicolumn{1}{c}{5.23}          & \multicolumn{1}{c}{8.11}          & 8.34           \\ \hline
                                   & 0.1                      & \multicolumn{1}{c}{39.42 $\pm$ 7.28}          & \multicolumn{1}{c}{33.49 $\pm$ 12.16}          & 20.42 $\pm$ 12.43          & \multicolumn{1}{c}{0.87}          & \multicolumn{1}{c}{0.74}          & 0.59          & \multicolumn{1}{c}{\textbf{0.09}} & \multicolumn{1}{c}{\textbf{0.15}} & 0.17          & \multicolumn{1}{c}{5.41}          & \multicolumn{1}{c}{12.52}         & 18.52          \\
                                   & 0.25                      & \multicolumn{1}{c}{\textbf{40.36 $\pm$ 8.07}} & \multicolumn{1}{c}{32.37 $\pm$ 12.02}          & 21.61 $\pm$ 12.88          & \multicolumn{1}{c}{\textbf{0.88}} & \multicolumn{1}{c}{0.72}          & 0.67          & \multicolumn{1}{c}{0.10}          & \multicolumn{1}{c}{0.17}          & 0.25          & \multicolumn{1}{c}{5.31}          & \multicolumn{1}{c}{10.23}         & 13.23          \\
IQN\cite{IQN2018}                               & 0.5                       & \multicolumn{1}{c}{38.47 $\pm$ 8.56}          & \multicolumn{1}{c}{29.90 $\pm$ 13.98}          & 19.76 $\pm$ 14.72          & \multicolumn{1}{c}{0.87}          & \multicolumn{1}{c}{0.73}          & 0.66          & \multicolumn{1}{c}{0.11}          & \multicolumn{1}{c}{0.19}          & 0.28          & \multicolumn{1}{c}{5.30}          & \multicolumn{1}{c}{8.46}          & 09.46          \\
                                   & 0.75                      & \multicolumn{1}{c}{37.62 $\pm$ 9.23}          & \multicolumn{1}{c}{25.72 $\pm$ 13.71}          & 18.32 $\pm$ 14.22          & \multicolumn{1}{c}{0.84}          & \multicolumn{1}{c}{0.69}          & 0.62          & \multicolumn{1}{c}{0.13}          & \multicolumn{1}{c}{0.22}          & 0.31          & \multicolumn{1}{c}{5.25}          & \multicolumn{1}{c}{8.57}          & 08.60          \\
                                   & 1.0                      & \multicolumn{1}{c}{38.70 $\pm$ 7.89}          & \multicolumn{1}{c}{23.39 $\pm$ 14.86}          & 16.29 $\pm$ 15.47          & \multicolumn{1}{c}{0.86}          & \multicolumn{1}{c}{0.67}          & 0.57          & \multicolumn{1}{c}{0.12}          & \multicolumn{1}{c}{0.30}          & 0.41          & \multicolumn{1}{c}{\textbf{5.12}} & \multicolumn{1}{c}{\textbf{7.89}} & \textbf{08.05} \\ \hline
ART-IQN                           & \multicolumn{1}{c|}{-}    & \multicolumn{1}{c}{39.85 $\pm$ 7.23}          & \multicolumn{1}{c}{\textbf{36.43 $\pm$ 13.51}} & \textbf{24.88 $\pm$ 12.31} & \multicolumn{1}{c}{0.87}          & \multicolumn{1}{c}{\textbf{0.77}} & \textbf{0.70} & \multicolumn{1}{c}{0.11}          & \multicolumn{1}{c}{0.17}          & \textbf{0.15} & \multicolumn{1}{c}{5.32}          & \multicolumn{1}{c}{8.26}          & 11.76          \\ \hline
\end{tabular}
\end{table*}


\subsubsection{Evaluation in Simulation}
To show the efficacy of our algorithm, ART-IQN is compared with IQN with multiple risk-tendencies $\alpha = \{0.1, 0.25, 0.5, 0.75, 1.0\}$. In addition, we also trained a DQN \cite{natureDQN2015Mnih} agent as our baseline following the same training procedure.
The average episodic return, success rate, collision rate and average navigation time are compared among agents across various environments. Specifically, the evaluation is executed on three sets of environments with $n_{obs} = \{2, 6, 12\}$ for each agent. Each set has $100$ diverse randomized environments.

\section{Results}
\label{sec:results}
\subsection{Simulation Results}

\subsubsection{Quantitative Analysis}
Table \ref{tab:eval} gives quantitative results by comparing: (1) DQN, (2) IQN with different risk tendencies and (3) ART-IQN. For all the environments, DQN performs similar to risk-neutral IQN. For $n_{obs}=2$, all the agents finish the navigation task at a high success rate and a low collision rate since the environment is comparatively easy. For $n_{obs}=6$, while IQN with lower CVaR values can maintain a lower collision rate compared to higher CVaR values, the task finishing time is longer and the timeout rate also increases.
\begin{figure}[!h]
    \centering
    \includegraphics[width=0.48\textwidth]{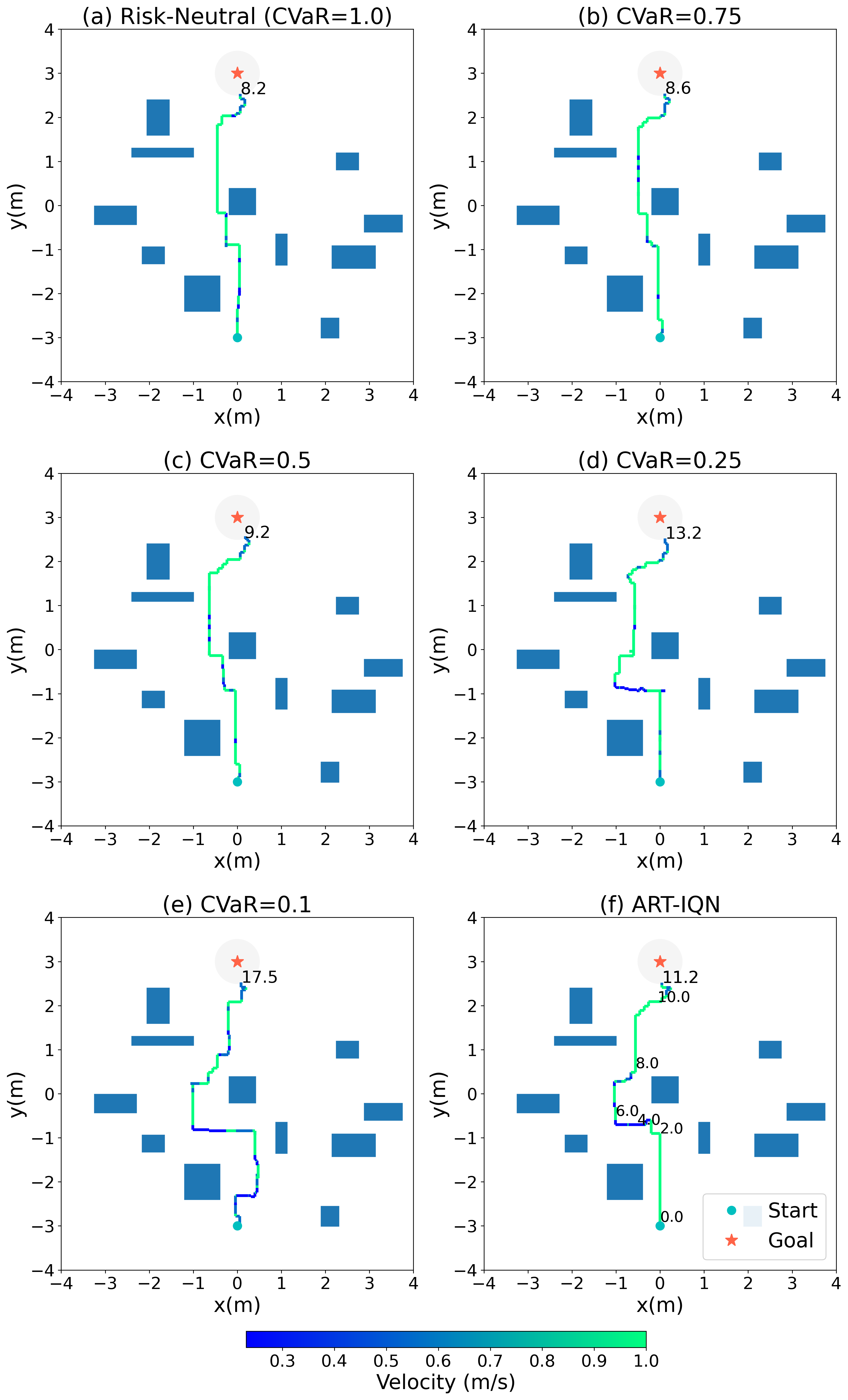}
    \caption{Drone behavior and navigation time comparison between agents with different risk-tendencies.}
    \label{fig:behavior}
\end{figure}
In contrast, ART-IQN achieves a success rate of $0.77$, maintains low collision rate and a average navigation time of only $0.37s$ longer compared to risk-neutral policy. For $n_{obs}=12$, the average return and success rate of all the agents decrease. This drop can be explained by the severe partial observability the agents face. While there are more cases of timeout with lower CVaR values and more collisions with a higher one, ART-IQN performs best in success rate and collision rate with a decent navigation time.

\subsubsection{Qualitative Analysis}


We demonstrate the behavior of agents in Fig. \ref{fig:behavior} by considering a typical environment encountered in evaluation. In Fig. \ref{fig:behavior} (a), risk-neutral IQN achieves the goal as fast as possible, ignoring the risk of getting too close to obstacles that leads to a higher rate of collision. On the other hand, risk-averse policies, especially the one with $\alpha=0.1$ as shown in Fig. \ref{fig:behavior} (e), generate safer policies but sacrifice in navigation efficiency. In contrast, as shown in Fig. \ref{fig:behavior} (f), ART-IQN acts adaptively - avoiding obstacles cautiously in the middle area where there is more uncertainty encountered, and flying at a higher speed when it's more certain about the current observation.

\begin{figure}[]
     \centering
    \includegraphics[width=0.48\textwidth]{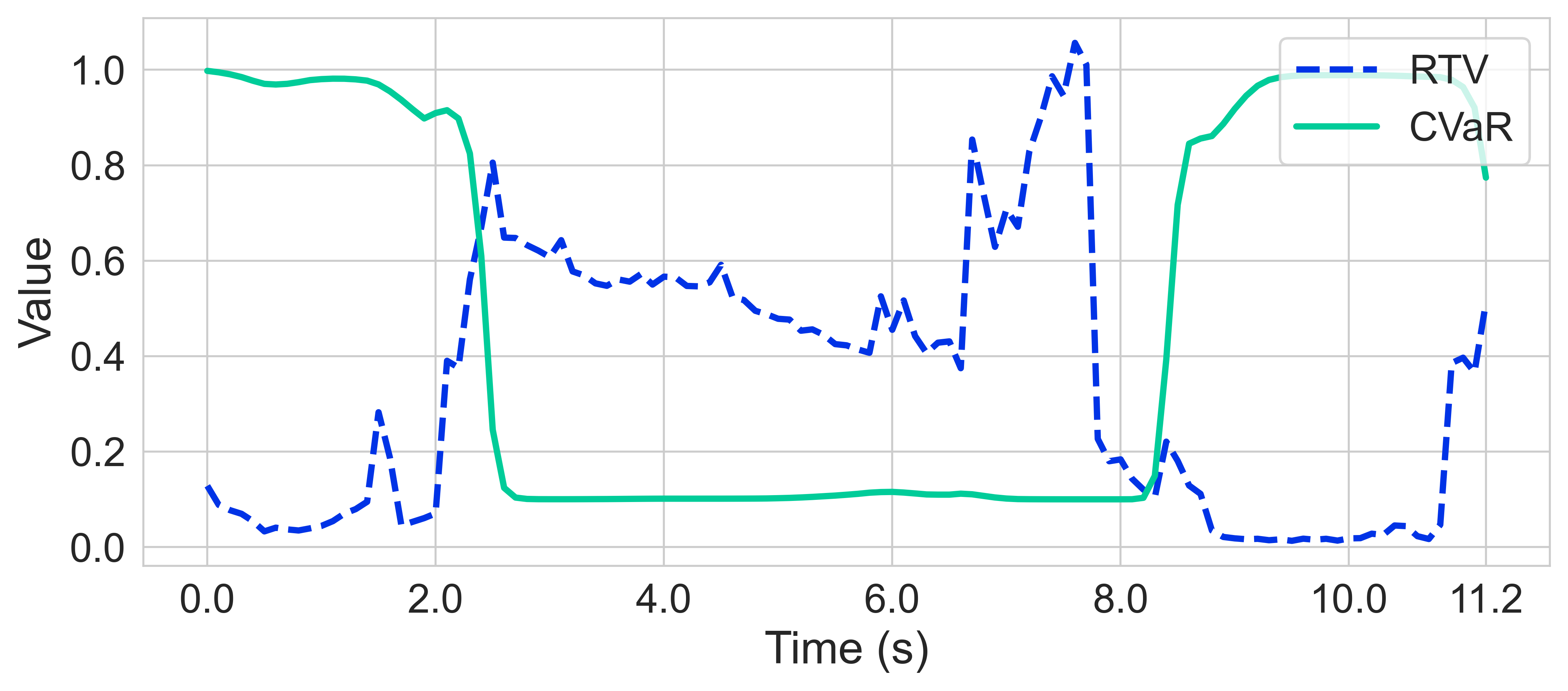}
     \caption{RTV and adaptive CVaR. ART-IQN can adapt its CVaR value accordingly with RTV as an estimation of intrinsic uncertainty in the environment.}
     \label{fig:tcv-cvar}
\end{figure}

Fig. \ref{fig:tcv-cvar} shows the RTV and adaptive CVaR along the trajectory in Fig. \ref{fig:behavior} (f). The drone starts with CVaR $\alpha=1.0$, which is risk-neutral since it does not know the environment yet. During $0s$ to $2s$, the drone flies at its maximum speed with risk-neutral policy as the current uncertainty is relatively low. Around $2s$ to $8s$, the intrinsic uncertainty estimated by RTV increases and stays at a high level, which resulted a decrease in CVaR values, corresponding to the risk-averse behavior in the middle area of Fig. \ref{fig:behavior} (f). When the RTV drops and stays at a low level from $8s$ until the end of the episode, the CVaR also decreases, resulting in a risk-neutral policy to reach the goal point efficiently. It is clear that, as the drone navigates through the environment, it can adjust its risk-tendency to be risk-averse when the uncertainty increases and to be risk-neutral when there is less uncertainty in the environment.

\begin{table*}[]
    \caption{\uppercase{Real-world experiment results}}
    \centering
    \label{tab:real}
\begin{tabular}{c|c|ccc|ccc}
\hline
                     & CVaR                      & \multicolumn{3}{c|}{\# Success / \# Collision}              & \multicolumn{3}{c}{Navigation time (mean $\pm$ std) $[s]$}                                        \\ \hline
Environment                  & -                         & \multicolumn{1}{c}{1}     & \multicolumn{1}{c}{2}     & {3} & \multicolumn{1}{c}{1}                & \multicolumn{1}{c}{2}                & {3} \\ \hline 
\multirow{2}{*}{IQN} & \multicolumn{1}{c|}{0.1} & \multicolumn{1}{c}{9 / 0} & \multicolumn{1}{c}{3 / 0} & \multicolumn{1}{c|}{3 / 0} & \multicolumn{1}{c}{14.41 $\pm$ 3.67} & \multicolumn{1}{c}{16.73 $\pm$ 0.49} & \multicolumn{1}{c}{21.52 $\pm$ 1.76} \\
                     & \multicolumn{1}{c|}{1.0} & \multicolumn{1}{c}{8 / 1} & \multicolumn{1}{c}{2 / 1} & \multicolumn{1}{c|}{0 / 3} & \multicolumn{1}{c}{10.12 $\pm$ 3.29} & \multicolumn{1}{c}{12.54 $\pm$ 0.58} & - \\ \hline
ART-IQN              & -                         & \multicolumn{1}{c}{9 / 0} & \multicolumn{1}{c}{3 / 0} & \multicolumn{1}{c|}{3 / 0} & \multicolumn{1}{c}{12.32 $\pm$ 3.46} & \multicolumn{1}{c}{13.37 $\pm$ 0.47} & \multicolumn{1}{c}{17.95 $\pm$ 1.85} \\ \hline
\end{tabular}
\end{table*}

\begin{figure*}[!h]
     \centering
     \begin{subfigure}[b]{0.32\textwidth}
         \centering
         \includegraphics[width=\textwidth]{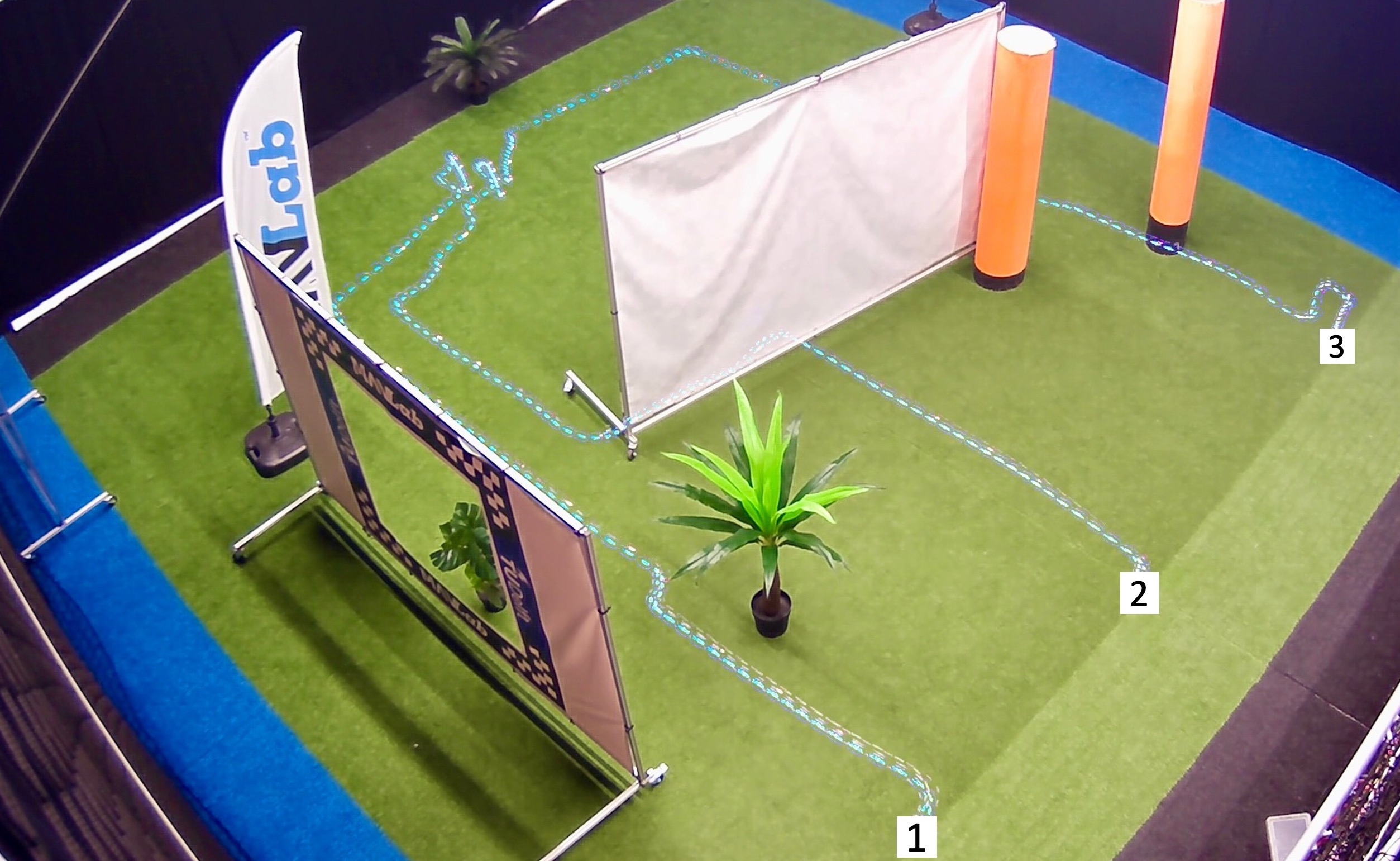}
         \caption{Environment 1: ART-IQN}
         \label{fig:traj_adaptive}
     \end{subfigure}
     \hfill
     \begin{subfigure}[b]{0.32\textwidth}
         \centering
         \includegraphics[width=\textwidth]{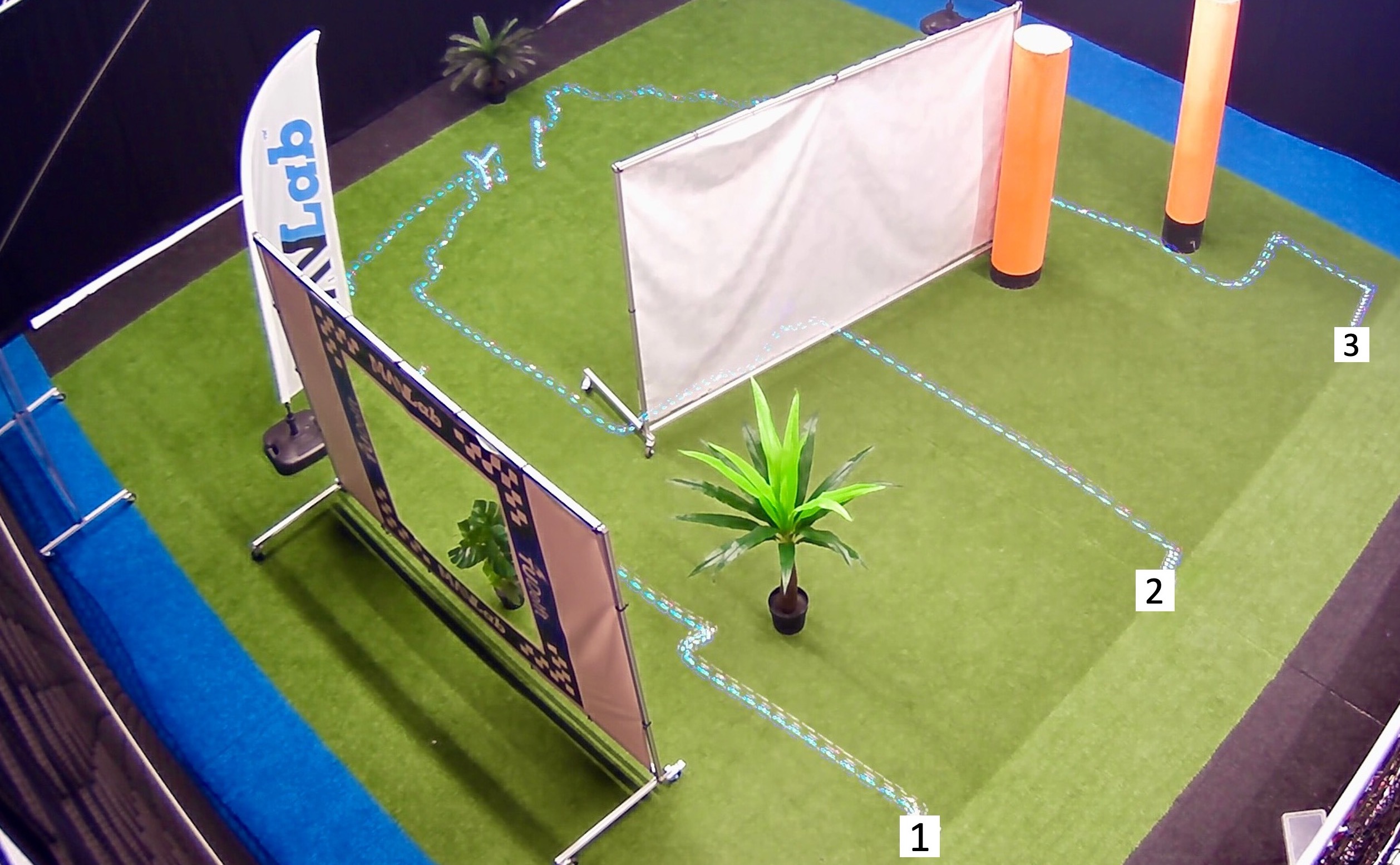}
         \caption{Environment 1: Risk-averse (CVaR=0.1)}
         \label{fig:traj_averse} 
     \end{subfigure}
     \hfill
     \begin{subfigure}[b]{0.32\textwidth}
         \centering
         \includegraphics[width=\textwidth]{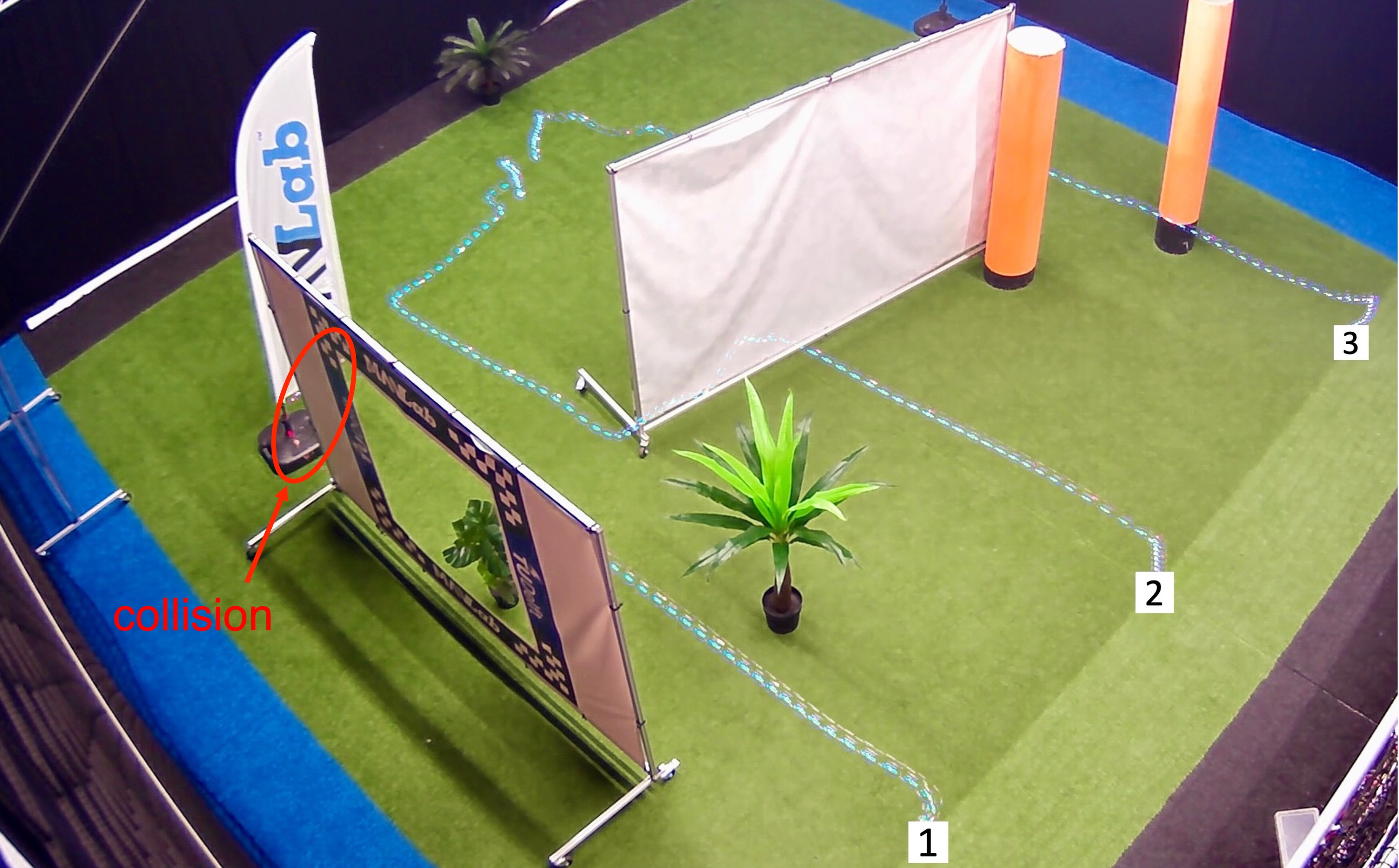}
         \caption{Environment 1: Risk-neutral}
         \label{fig:traj_neutral}
     \end{subfigure}
     \hfill
     \begin{subfigure}[b]{0.32\textwidth}
         \centering
         \includegraphics[width=\textwidth]{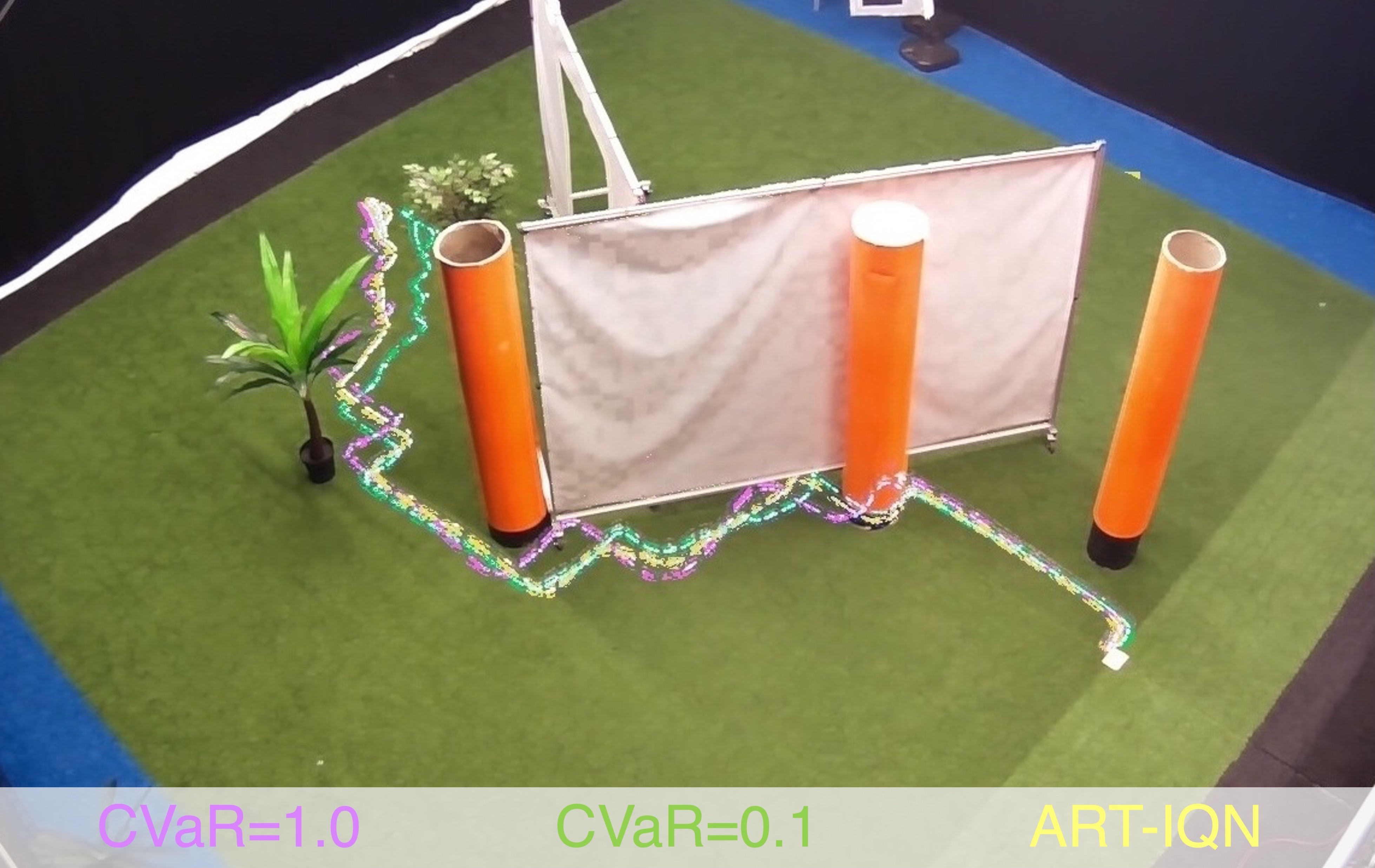}
         \caption{Environment 2}
         \label{fig:traj_addition}
     \end{subfigure}
     \hfill
     \begin{subfigure}[b]{0.32\textwidth}
         \centering
         \includegraphics[width=\textwidth]{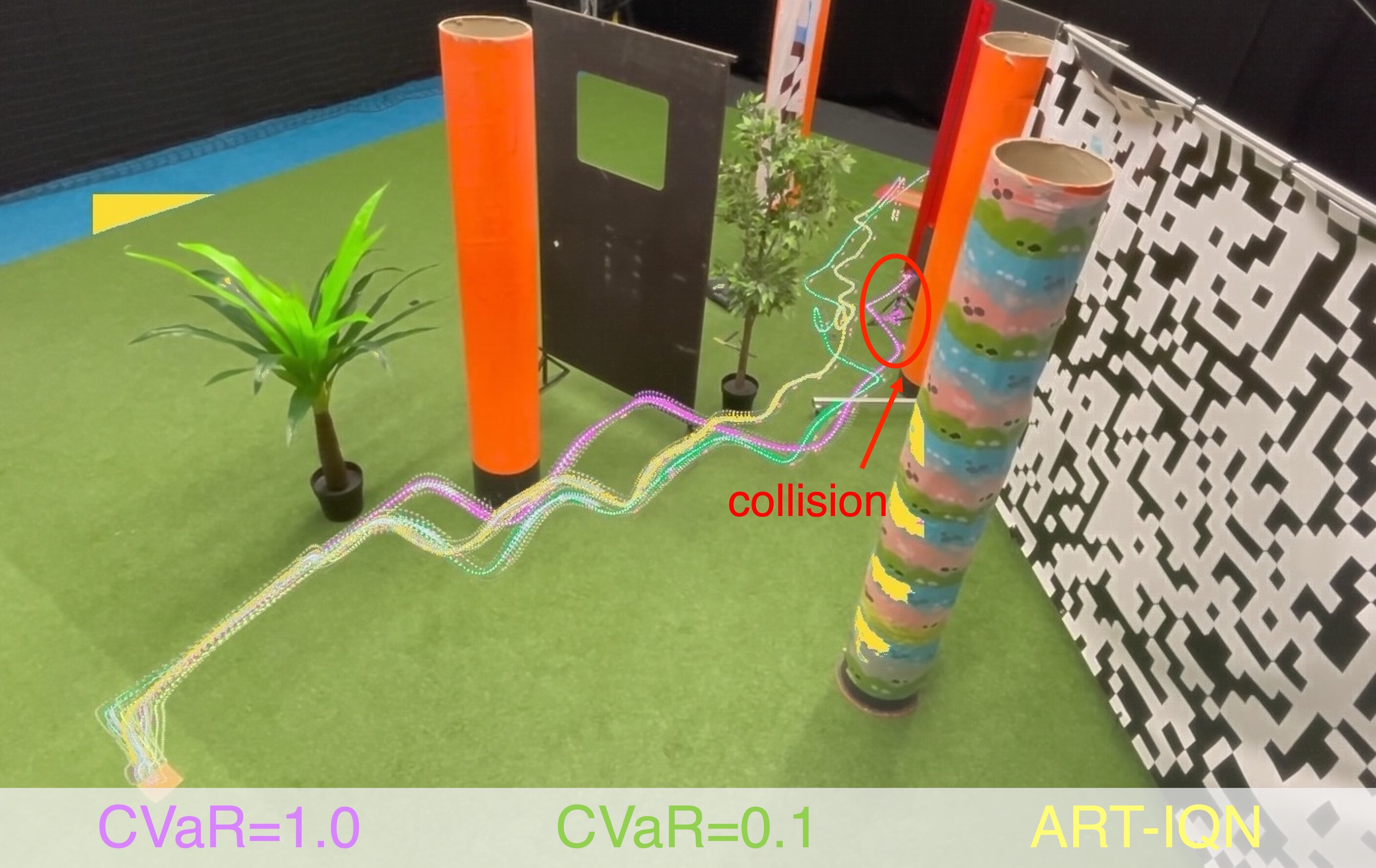}
         \caption{Environment 3}
         \label{fig:traj_env3}
     \end{subfigure}
     \hfill
     \begin{subfigure}[b]{0.32\textwidth}
         \centering
         \includegraphics[width=\textwidth]{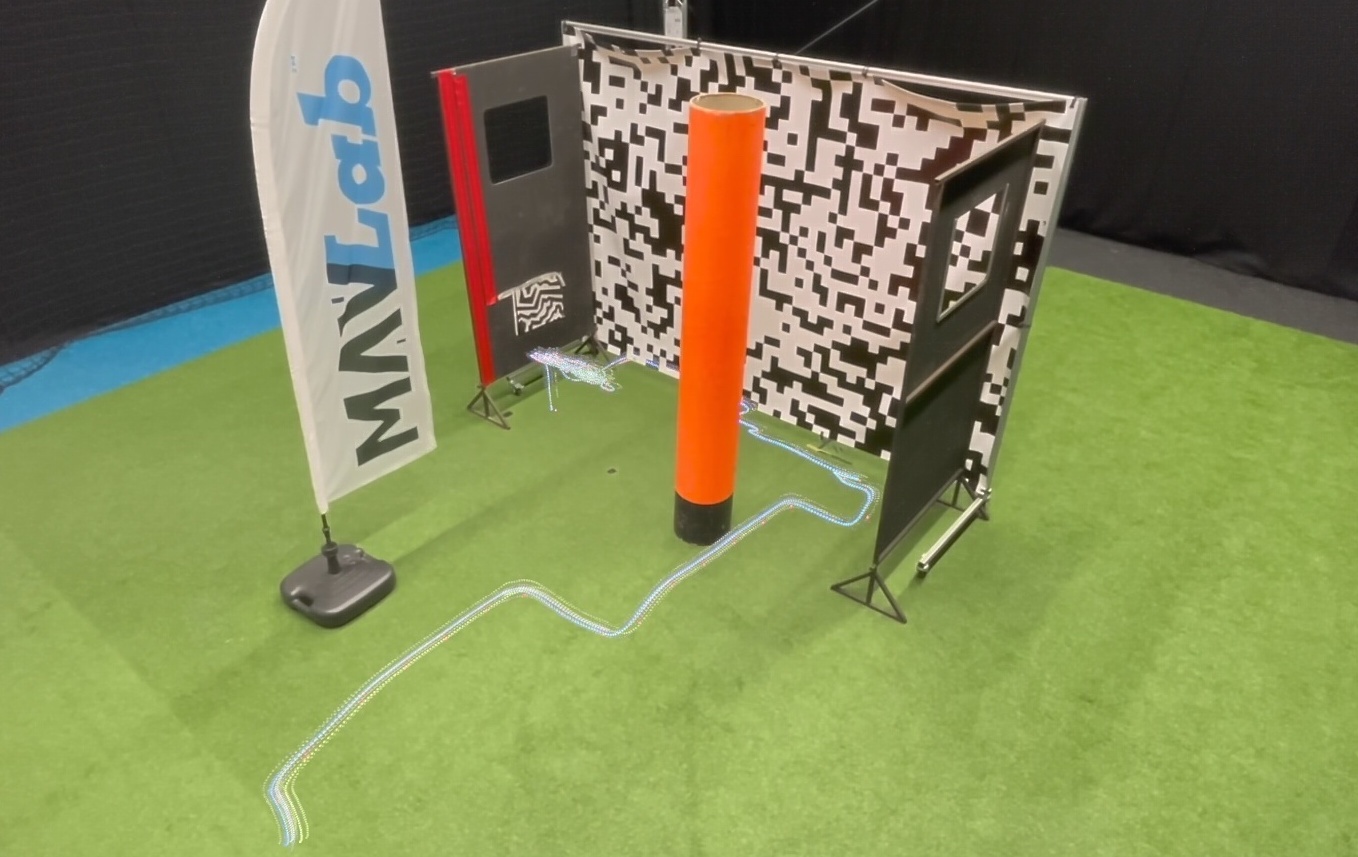}
         \caption{Adversarial environment: ART-IQN}
         \label{fig:traj_adversarial}
     \end{subfigure}

     \caption{Image frames for real-world experiments. The trajectory is the blue LED on the drone. For environment 2 and 3, the trajectories generated by different agents are distinguished with recolored LEDs.}
     \label{fig:cyberzoo}
\end{figure*}

\subsection{Real-World Experiments}
\subsubsection{Hardware Setup}
The Crazyflie nano drone we used for real-world experiments is shown in Fig. \ref{fig:crazyflie}. It has dimensions $92\times92\times29mm$ and weighs $27.5 g$. Policy is performed on a laptop, which gives velocity command and communicates with Crazyflie via a radio-to-USB dongle. The velocity command period $T$ is set to be the same as in simulation. The drone navigation task is implemented in $3$ different $10 \times 10 m$ cluttered environments. As shown in Fig. \ref{fig:cyberzoo}, we put artificial trees, boards and cylinders as obstacles in the environment that the agent has not seen in the simulation. Reflective markers are attached on four propeller hubs to let the motion capturing system track the global position of the drone.

\subsubsection{Evaluation Results}
We test IQN with $\alpha=\{0.1, 1.0\}$ and ART-IQN for comparison. In environment 1, each agent is initialized at $3$ different starting points for $3$ runs, resulted a total number of $27$ runs. In environment 2 and a more complex environment 3, the drone takes off at the same starting point. As shown in Table \ref{tab:real}, all runs succeeded except risk-neutral policies. Although risk-neutral IQN achieves fastest navigation in succeeded runs, it ignores the risk in the environment causing the drone to collide with obstacles. Risk-averse IQN succeeds in all experiments without collisions, but with a loss of navigation efficiency. In contrast, ART-IQN navigates through all the environments safely and efficiently.

Fig. \ref{fig:cyberzoo} (a)-(e) demonstrate diverse behaviors among different risk-tendencies. Unlike risk-neutral IQN, both ART-IQN and risk-averse IQN keep a safe distance to obstacles to avoid collisions. The advantage of ART-IQN compared to risk-averse IQN is mainly reflected in the shorter navigation time as in Table \ref{tab:real}. Additionally, we also designed a U-shape adversarial environment to study the generalization capability of IQN agents. However, all agents including ART-IQN stick around in the corner of the U-shape obstacle as shown in Fig. \ref{fig:cyberzoo} (f). It is highly possible that U-shape obstacles have not been seen often by the agent and can be solved by generating similar situations in the training environment.

\section{Conclusion}
In conclusion, focusing on the autonomous drone navigation under partial observability, we propose an adaptive risk-tendency algorithm based on distributional RL to adapt risk-tendency according to the estimated intrinsic uncertainty. Our algorithm uses EWAF to adjust risk-tendency represented by the CVaR, with lower tail conditional variance as an estimation of the intrinsic uncertainty. We show the effectiveness of our algorithm both in simulation and real-world experiments. Empirical results show that our algorithm can adaptively balance the efficiency-safety trade-off.


However, the step size $\eta$ is currently pre-specified, it would be worthy to optimize it. Despite that, we believe our method could serve as a first step to develop risk-tendency adaptation methodologies for distributional RL applications especially in risk-sensitive settings.



\section*{Acknowledgment}
The authors would like to thank Jinke He for the discussions and Bart Duisterhof, Yingfu Xu for the real-world experiment setup.

\bibliographystyle{./style/IEEEtran}
\bibliography{./style/IEEEabrv, ./iros2022}

\addtolength{\textheight}{-10cm}
\end{document}